\documentclass[11pt]{article}

\usepackage[preprint]{acl}

\usepackage{times}
\usepackage{latexsym}

\usepackage[T1]{fontenc}

\usepackage[utf8]{inputenc}

\usepackage{microtype}

\usepackage{inconsolata}

\usepackage{graphicx}
\usepackage{hyperref}
\usepackage{url}
\usepackage{booktabs}
\usepackage{amsmath}
\usepackage{wrapfig}
\usepackage{tabularx}
\usepackage{caption}
\usepackage{amssymb}

\title{GRACE: Graph-Grounded Reflective Agent Copilot Engine for Expert-in-the-Loop Knowledge Expansion}

\author{
  \textbf{John Seon Keun Yi}$^{1}$,
  \textbf{Joshua R. Minot}$^{2}$,
  \textbf{Dokyun Lee}$^{1}$
  \\
  $^{1}$Boston University \quad $^{2}$MassMutual
  \\
  \texttt{\{jskyi, dokyun\}@bu.edu}, \quad \texttt{jminot26@massmutual.com}
}

\begin{document}
\maketitle
\begin{abstract}
Large language models deployed in high-stakes settings frequently generate plausible but ungrounded claims. Standard retrieval-augmented generation (RAG) pipelines offer limited remedy, since they retrieve isolated passages without tracking cross-document evidence relationships or quantifying uncertainty. 
We introduce GRACE (Graph-grounded Reflective Agent Copilot Engine), a framework that deconstructs LLM responses into atomic claims and grounds them against trusted knowledge priors within a weighted bipartite graph. 
Edge weights encode the closeness of each claim to the priors, enabling weighted centrality analysis that classifies claims as Grounded, Refuted, or Boundary. 
Such classification identifies not just hallucinations but also novel or contested claims at the frontier of the model's knowledge.
To efficiently allocate human or agent resources, we formulate a Return on Attention (RoA) objective that defers a claim to expert review only when its priority-weighted uncertainty exceeds the cost of verification. 
Claims verified by experts are promoted to new evidence anchors, closing a validator-LLM evolutionary loop that expands the knowledge base across iterations. 
We evaluate GRACE across multiple language models and on datasets spanning both general and domain-specific knowledge. 
Our results show that our knowledge base serves as a reliable foundation for retrieval that outperforms RAG baselines, and that the RoA framework efficiently selects valuable boundary knowledge for expert verification. 
These findings demonstrate that graph-structured representations combined with expert-in-the-loop verification can mitigate hallucination at the system level rather than at the generation level. Code available at \url{https://github.com/johnsk95/grace_code}
\end{abstract}

\section{Introduction}
Large language models (LLMs) have driven widespread adoption across industries, with recent analyses reporting that over 78\% now regularly use generative AI in at least one business function~\cite{singla2025state}. Despite this surge in adoption, scaled deployment remains low. More than 90\% of function-specific use cases remain stuck in the pilot stage~\cite{mckinsey2024genai}.
The primary bottleneck is reliability: inaccuracy is consistently cited as the top risk in deploying generative AI systems~\cite{sakib2024risks, sukharevsky2025seizing}, preventing their use in high-stakes decision-making where precision is non-negotiable.
When prompted with out-of-knowledge queries, LLMs confabulate rather than abstain, producing fluent, confident-sounding outputs that are factually ungrounded. Any system that relies solely on LLM-internal consistency to judge truthfulness cannot distinguish between well-supported facts and shared hallucinations~\cite{lin2022truthfulqa, xu2024hallucination}: false claims that are confidently and repeatedly generated across multiple samples.

Current mitigation strategies are insufficient to solve this problem. 
Retrieval-augmented generation (RAG)~\cite{lewis2020retrieval} retrieves isolated document chunks to condition generation, but cannot link patterns across documents or track how evidence relates to specific claims. 
Uncertainty estimation methods~\cite{jiang2024graph, xia2025survey} offer complementary signals but still rely on the capacity of the base language model to judge its own reliability. 
None of these approaches provides the full auditability that high-stakes enterprise deployment demands.
We argue that grounding LLM outputs against validated external \textit{knowledge priors}, combined with structured expert verification, provides the necessary signal for safe and scalable deployment.

We introduce GRACE (Graph-grounded Reflective Agent Copilot Engine), a framework that decomposes LLM responses into atomic claims and grounds them within a weighted bipartite graph alongside trusted knowledge priors. 
Edge weights encode the closeness of each claim to these priors, and weighted centrality analysis classifies claims into three categories: Grounded (high consensus and anchored to priors), Refuted (contradicting established knowledge), and Boundary (novel, contested, or sparsely connected).
To convert this map into a growing knowledge base, GRACE formulates a Return on Attention (RoA) objective that routes only the most valuable boundary claims to experts for verification.
Verified claims are promoted to new priors, closing a co-evolutionary loop that iteratively expands the knowledge base with each round of expert feedback. 
We evaluate GRACE across multiple language models and on datasets spanning both general and domain-specific knowledge. Our experiments demonstrate that: (1) the co-evolved knowledge base serves as a reliable foundation for downstream retrieval, outperforming RAG-based baselines; (2) the RoA framework efficiently selects high-value boundary claims for verification and knowledge expansion; and (3) displays robustness in scaled settings.

\section{Related Work}
\paragraph{Retrieval-Augmented Generation.}
RAG \citep{lewis2020retrieval} established the paradigm of conditioning LLM generation on retrieved passages, combining parametric and non-parametric memory for knowledge-intensive tasks. Self-RAG \citep{asai2023self} extends this with adaptive retrieval and self-critique, while GraphRAG \citep{han2024retrieval, edge2024local} constructs knowledge graphs and community summaries to support query-focused summarization over broader contexts. Despite these advances, existing RAG systems retrieve isolated passages or subgraph summaries without tracking how individual claims relate to trusted evidence sources or quantifying claim-level uncertainty.

\paragraph{Uncertainty Quantification in LLMs.}
Methods for quantifying LLM reliability span token-level entropy \citep{ye2024benchmarking}, semantic entropy over meaning-clustered generations \citep{farquhar2024detecting}, and self-consistency checks across sampled responses \citep{manakul2023selfcheckgpt}. Most directly related to our work, \citet{jiang2024graph} model response-claim relationships as a bipartite graph and use centrality metrics to estimate claim-level uncertainty. However, all approaches rely on LLM-internal consistency and therefore cannot separate genuine facts from shared hallucinations. GRACE extends the graph-based framework of \citet{jiang2024graph} by introducing external knowledge priors as anchors, shifting the basis of uncertainty from internal consistency to external grounding, and coupling it with a expert-in-the-loop expansion mechanism.

\section{Method} \label{sec:method}
\begin{figure*}
    \centering
    \includegraphics[width=1\linewidth]{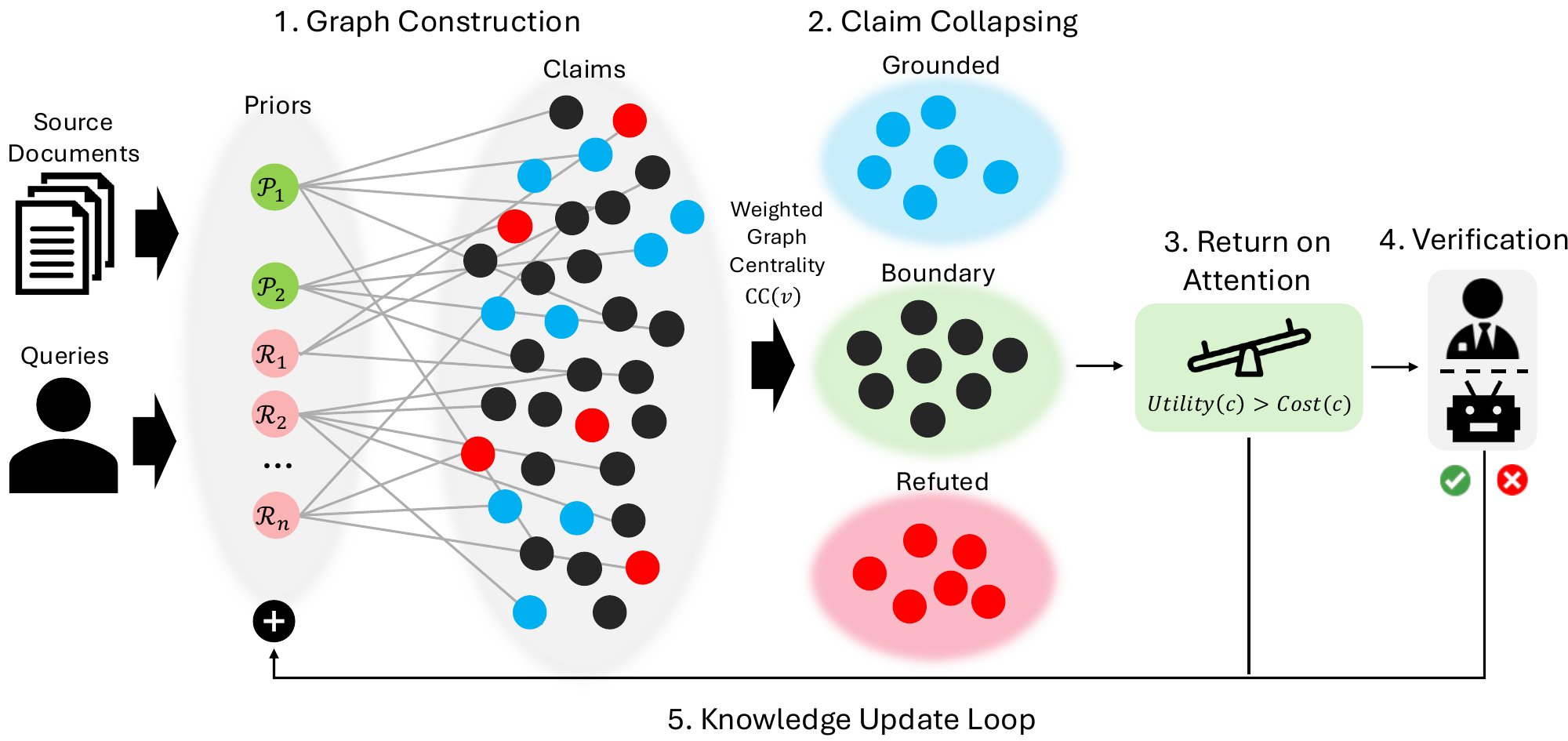}
    \caption{\textbf{Overview of the GRACE framework.} Given user queries and source documents, priors $\mathcal{P}$ and LLM responses $\mathcal{R}$ are decomposed into atomic claims and organized into a weighted bipartite graph. Closeness centrality $CC(v)$ classifies each claim as Grounded, Boundary, or Refuted. The Return on Attention (RoA) framework routes only boundary claims whose expected utility exceeds verification cost to experts (human or LLM agent). Confirmed claims are promoted to new priors, closing one knowledge expansion loop.}
    \label{fig:teaser}
\end{figure*}

Given user queries and a set of data sources, GRACE produces claims with auditable status labels (Grounded, Refuted, Boundary) and iteratively expands a trusted knowledge base through targeted expert verification. Figure~\ref{fig:teaser} provides an overview of the full pipeline.

\subsection{Preliminaries: Graph-Based Uncertainty Estimation}
GRACE builds on the graph-based uncertainty framework of \citet{jiang2024graph}, which models the relationship between sampled responses and their constituent claims as a bipartite graph. Given a query, multiple responses $R = \{r_1, r_2, \ldots\}$ are sampled from an LLM, and each response is decomposed into atomic, sentence-level claims $C = \{c_1, c_2, \ldots\}$. A bipartite graph $G = (R \cup C, E)$ is constructed where an edge $(r_i, c_j) \in E$ exists if response $r_i$ entails claim $c_j$. Graph centrality metrics (e.g., closeness centrality) then estimate each claim's uncertainty, since ``true'' claims tend to be more central in the graph, supported by many independent responses. However, since this approach relies entirely on LLM-internal consistency, it cannot distinguish genuine facts from shared hallucinations, i.e., false claims that receive high centrality simply because they are repeatedly generated. 

\subsection{Bipartite Graph Construction}
To overcome this limitation, we introduce \textit{knowledge priors}: trusted external information sources such as verified documents, compliance rulebooks, standard operating procedures, or expert-confirmed records. We integrate these priors directly into the graph structure to serve as anchors of truth, enabling the system to distinguish genuinely grounded claims from shared hallucinations.
 
\paragraph{Graph structure.} We construct a weighted bipartite graph $G = ((R \cup P, C), E)$ where $R = \{r_1, r_2, \ldots\}$ is the set of sampled LLM responses, $P = \{p_0, p_1, \ldots\}$ is the set of trusted priors, and $C$ is the set of unique atomic claims decomposed from both $R$ and $P$. An edge $e = (u, v) \in E$ exists if a source node $u \in (R \cup P)$ entails a claim node $v \in C$. Let $C_P \subseteq C$ denote the subset of claims that are entailed by at least one prior.
 
\paragraph{Edge weighting.} Unlike the unweighted baseline, we define a weighting function $W: E \to \mathbb{R}^+$ that encodes closeness to priors. Edges are assigned weights according to their relationship type:

\begin{equation}
\label{eq:weights}
w_{uv} = \begin{cases}
w_{\text{anchor}} & \text{if } u \in P \text{ and } v \in C_P \quad  \\
w_{\text{agree}} & \text{if } u \in R \text{ and } v \in C_P \quad  \\
w_{\text{neutral}} & \text{if } u \in R \text{ and } v \notin C_P \quad  \\
w_{\text{contra}} & \text{if } v \text{ contradicts } P \quad 
\end{cases}
\end{equation}

where $w_{\text{anchor}} > w_{\text{agree}} > w_{\text{neutral}} > w_{\text{contra}}$. We define the traversal cost of an edge as $\text{cost}(u,v) = 1 / w_{uv}$, so that paths through priors have the lowest cost. This weighting hierarchy ensures that centrality metrics naturally favor claims that are not merely consistent across LLM responses but actively grounded in verified knowledge.

\subsection{Scoring and Knowledge Boundary Classification}
With the weighted graph in place, GRACE computes centrality-based scores to classify each claim and map the LLM's knowledge boundary.
 
\paragraph{Grounded closeness centrality.} We adopt closeness centrality as the primary scoring metric, following the findings of~\citet{jiang2024graph} that it is highly effective for detecting false claims. To handle potential graph disconnectedness, we use the modified closeness centrality variant adapted for our weighted structure:

\begin{equation}
\label{eq:cc}
CC(v) = \frac{N - 1}{\sum_{u \in V} d_W(v, u)} \cdot \frac{|V_v|}{N}
\end{equation}
where $N$ is the total number of nodes, $|V_v|$ is the size of the connected component containing $v$, and $d_W(v, u)$ is the shortest-path distance between $v$ and $u$ computed using edge costs $1/w_{uv}$ via Dijkstra's algorithm. This metric assigns higher centrality to claims reachable via low-cost Anchor and Agreement links, effectively boosting grounded facts while pushing out ungrounded hallucinations.

\paragraph{Boundary classification.} Using the closeness centrality scores, GRACE classifies each claim into one of three statuses:
\textbf{Grounded}: Claims with high centrality scores that are strongly connected to priors through Anchor and Agreement links. These represent verified, trustworthy knowledge.
\textbf{Refuted}: Claims with very low centrality scores, connected primarily through Contradictory links. These are identified hallucinations and are archived or discarded.
\textbf{Boundary}: Claims with intermediate centrality scores that are neither well-supported by priors nor clearly contradicted. These include \textit{novel} claims (potentially true but unverified information outside existing priors) and \textit{contested} claims (ambiguous or low-consensus). Boundary claims define the frontier of the LLM's knowledge and are candidates for verification.

Formally, we define the set of boundary claims as $C_{\text{boundary}} = C \setminus C_P$, or equivalently as claims below a centrality threshold $\delta$: $C_{\text{boundary}} = \{c \in C \mid CC(c) < \delta\}$.

\subsection{Return on Attention: Validator Deferral and Knowledge Expansion}
Identifying the knowledge boundary is only the first step. GRACE closes the loop by routing the most valuable boundary claims to expert validators and promoting verified claims to new priors. We formalize this as the \textit{Return on Attention} (RoA) framework.
 
\paragraph{Priority scoring.} For each boundary claim $c \in C_{\text{boundary}}$, we compute a priority score as a weighted combination of two complementary signals:
\begin{equation}
\label{eq:priority}
\text{Priority}(c) = \lambda_c \cdot S_{\text{centrality}}(c) + \lambda_n \cdot S_{\text{novelty}}(c)
\end{equation}
where $\lambda_c, \lambda_n$ are tunable hyperparameters that balance the selection strategy.
 
The \textit{centrality score} identifies ``source'' claims inside the unresolved subgraph whose verification would cascade to resolve clusters of related claims. Let $G_{\text{boundary}} = ((R, C_{\text{boundary}}), E')$ be the subgraph consisting only of LLM responses and boundary claims. We compute the betweenness centrality $C_B(c, G_{\text{boundary}})$ for each claim $c$ in this subgraph:

\begin{equation}
\label{eq:centrality}
S_{\text{centrality}}(c) = \frac{C_B(c, G_{\text{boundary}})}{\max_{c' \in C_{\text{boundary}}} C_B(c', G_{\text{boundary}})}
\end{equation}
The \textit{novelty score} identifies claims that are most distant from any established knowledge, representing frontiers for knowledge expansion. Using the weighted shortest-path distance, we compute:
\begin{equation}
\label{eq:novelty}
\begin{split}
    d_{\min}(c) = \min_{c^* \in C_P} d_W(c, c^*) \\
\quad S_{\text{novelty}}(c) = \frac{d_{\min}(c)}{\max_{c' \in C_{\text{boundary}}} d_{\min}(c')}
\end{split}
\end{equation}
 
\paragraph{Deferral objective.} For each claim, the system chooses between two resolution paths: (1) \textit{auto-resolve} (upper line in the equation below) by accepting the graph-based classification, or (2) \textit{defer} (lower line in equation) to an expert at some cost. The utility of each claim is:
\begin{equation}
\label{eq:utility}
\text{Utility}(c) = \max \begin{cases}
\sigma(CC(c)) \times \text{Priority}(c) \\
\text{Priority}(c) - \text{Cost}(c)
\end{cases}
\end{equation}
where $\sigma(CC(c))$ is the grounded closeness centrality normalized via a sigmoid function, representing the system's confidence in its own classification, and $\text{Cost}(c)$ is the cost of expert review. The system selects expert verification if and only if:
\begin{equation}
\label{eq:deferral}
\text{Priority}(c) \cdot \big(1 - \sigma(CC(c))\big) > \text{Cost}(c)
\end{equation}
This formalizes the RoA principle: validator effort is invested only when the claim is both important (high priority) and uncertain (low confidence), and the expected gain exceeds the cost of intervention.
Given a limited verification budget $B$, the system selects a subset $\mathcal{D} \subseteq C_{\text{boundary}}$ of claims to defer such that $\sum_{c \in \mathcal{D}} \text{Cost}(c) \leq B$. A greedy selection strategy ranks candidates by their cost-adjusted marginal utility:
\begin{equation}
\label{eq:selection}
\text{Score}(c) = \frac{\text{Priority}(c) \cdot \big(1 - \sigma(CC(c))\big)}{\text{Cost}(c)}
\end{equation}
and selects in decreasing order until the budget is exhausted. High-centrality nodes act as sources that can collapse clusters of related claims, while high-novelty nodes expand the knowledge frontier into previously unexplored regions.
 
\paragraph{Knowledge update loop.} Expert validator review deferred claims and decide to either confirm or reject. Confirmed claims are promoted to new priors $P' = P \cup \{c_{\text{verified}}\}$, and edge weights throughout the graph are recomputed to reflect the expanded knowledge base. Rejected claims are archived as known hallucinations. Over successive iterations, this co-evolutionary loop accumulates the knowledge base: each round of expert feedback expands the set of priors, which in turn enables the graph to resolve more claims autonomously in subsequent iterations, progressively reducing the need for expert intervention.

\section{Experiments}
In this section, we detail a set of experiments to evaluate the performance of our GRACE framework. Section ~\ref{sec:retrieval} tests its retrieval ability, section~\ref{sec:expansion} validates knowledge expansion through oracle feedback, and section~\ref{sec:scalability} tests scalability to a large set of documents.

\subsection{GRACE as a Retrieval Tool}\label{sec:retrieval}
We design this experiment to answer the research question: \textit{Can GRACE serve as a reliable knowledge base for retrieval?}
Unlike traditional retrieval methods that rely solely on surface-level similarity between a query and a document corpus, GRACE constructs a structured knowledge graph where every claim is scored by its closeness centrality to trusted priors. This experiment tests whether that graph structure can surface the most relevant and trustworthy passages when presented with a downstream query, effectively functioning as a grounded retrieval engine.

\paragraph{Datasets.} We evaluate on two QA benchmarks containing long-form documents as context, replicating real-world scenarios where reference documents are lengthy and complex.
QASPER~\cite{dasigi2021dataset} is a dataset of NLP research papers paired with free-form questions. Each paper is divided into named sections; we use the Title+Abstract, Introduction, and Methods sections as the three priors for graph construction. We randomly sample 100 papers from the test split.
QuALITY~\cite{pang2022quality} is a multiple-choice QA dataset over long documents (articles, stories, essays) averaging roughly 5,000 tokens in length. Since these documents lack explicit section boundaries, we segment each article into sequential chunks of approximately 1{,}000 words, each serving as a prior. We randomly select 50 articles from the validation set.

\paragraph{Models.}
We test with two language models as the QA backbone to measure adaptability across model scales: GPT-4o-mini~\cite{gpt4o-mini}, a large cloud-based model, and LLaMA~3.1 8B~\cite{dubey2024llama}, a small locally-hosted model. For QASPER, open-ended answers are judged for correctness by GPT-4.1; for QuALITY, multiple-choice accuracy is computed via exact match against the gold label.

\paragraph{Baselines.}
We compare GRACE against two zero-shot conditions and two retrieval-based methods.
\textbf{Zero-shot + Part} provides only a partial document context to the LLM: the Title+Abstract section for QASPER, and the first chunk for QuALITY. This represents the minimal-context baseline.
\textbf{Zero-shot + Full} passes the full document as context. This serves as an upper bound on available information.
\textbf{RAG}~\cite{lewis2020retrieval} uses a standard retrieval-augmented generation pipeline: passages are retrieved by embedding similarity and top-$k$ results are provided as context.
\textbf{GraphRAG}~\cite{han2024retrieval} constructs a knowledge graph and community summaries from the document, then retrieves from these summaries to answer queries.

\paragraph{Metrics.}
We report two metrics: \textbf{Accuracy}, the percentage of questions answered correctly, and \textbf{Answerability}, the percentage of questions for which the model attempts an answer rather than abstaining. Answerability captures the model's calibration---an ideal system should abstain when its retrieved context is insufficient, rather than hallucinating an answer. We omit answerability for the QuALITY benchmark since all methods end up providing a ``best answer'' among the multiple choice options even with limited context. 

\paragraph{GRACE Retrieval.}
For GRACE, we first construct the weighted bipartite graph for each document as described in Section~\ref{sec:method}. At query time, source passages are scored using a combined retrieval metric:
\begin{equation}\label{eq:retrieval_score}
    \text{score}(q, p) = \alpha \cdot \text{sim}(\mathbf{e}_q, \mathbf{e}_p) + (1 - \alpha) \cdot CC(p)
\end{equation}
where $\text{sim}(\mathbf{e}_q, \mathbf{e}_p)$ is the cosine similarity between the question and passage embeddings, $CC(p)$ is the normalized closeness centrality of the passage's associated claims, and $\alpha = 0.5$. Passages are deduplicated by source prior and the top-5 are provided as context for question answering. This scoring mechanism ensures that retrieved passages are not only semantically relevant to the query but also well-grounded in the knowledge graph.

\begin{table*}[t]
\centering
\small
\begin{tabularx}{\textwidth}{@{}l@{\hspace{4pt}}@{\hspace{8pt}}>{\centering\arraybackslash}X@{\hspace{2pt}}>{\centering\arraybackslash}X@{\hspace{8pt}}>{\centering\arraybackslash}X@{}}
\toprule
& \multicolumn{2}{c}{QASPER [\cite{dasigi2021dataset}]} & {QuALITY [\cite{pang2022quality}]} \\
\cmidrule(lr){2-3}\cmidrule(lr){4-4}
& Accuracy (\%) & Answerable (\%) & Accuracy (\%) \\
\noalign{\vskip 2pt}\hline\noalign{\vskip 2pt}
\multicolumn{4}{c}{\textbf{GPT-4o-mini} [\cite{gpt4o-mini}]} \\
\noalign{\vskip 2pt}\hline\noalign{\vskip 2pt}
Zero-shot + Part   & 10.8 & 26.6 & 56.2 \\
Zero-shot + Full & \underline{29.0} & 53.6 & \underline{79.6} \\
RAG                    & 19.8 & 45.2 & 66.5 \\
GraphRAG               & 27.8 & 90.4 & 85.3 \\
\textbf{GRACE} (Ours)  & \textbf{28.1} & \textbf{50.6} & \textbf{77.0} \\
\noalign{\vskip 2pt}\hline\noalign{\vskip 2pt}
\multicolumn{4}{c}{\textbf{LLaMA 3.1 8B} [\cite{dubey2024llama}]} \\
\noalign{\vskip 2pt}\hline\noalign{\vskip 2pt}
Zero-shot + Part   & 10.7 & 26.6 & 41.1 \\
Zero-shot + Full & \underline{29.0} & 29.0 & \underline{45.1} \\
RAG                    & 18.5 & 35.6 & 58.9 \\
GraphRAG               & 29.0 & 95.2 & 29.1 \\
\textbf{GRACE} (Ours)  & \textbf{28.1} & \textbf{50.5} & \textbf{44.2} \\
\bottomrule
\end{tabularx}
\caption{\textbf{GRACE serves as a reliable retrieval tool over other retrieval and zero-shot methods.} Numbers indicate average accuracy (\%) and percentage of questions answerable (\%) across two QA benchmarks. GRACE displays competitive QA performance, while not being overly confident in question answerability. Underlined numbers indicate the full context ceiling.}
\label{tab:retrieval}
\end{table*}

\textbf{Results.} Table~\ref{tab:retrieval} reports accuracy and answerability across both benchmarks. On QASPER, GRACE matches the full-context zero-shot baseline on both models while retrieving only a small subset of the document, and substantially outperforms standard RAG. GraphRAG achieves comparable accuracy but reports answerability above 90\% on both models, indicating substantial overconfidence. GRACE, by contrast, reports answerability closely aligned with its accuracy, demonstrating that closeness centrality provides a natural calibration mechanism. On QuALITY, GRACE performs within a few points of the full-context baseline with GPT-4o-mini, while GraphRAG collapses to 29.1\% on LLaMA despite leading on the larger model. Notably, GRACE's accuracy is robust to model scale: on QASPER it maintains nearly identical performance across GPT-4o-mini and LLaMA, whereas every other method degrades. This suggests that smaller models benefit most from a pre-structured, grounded retrieval source that reduces the reasoning burden.

\subsection{Knowledge Expansion}\label{sec:expansion}
This experiment is aimed at answering the question: \textit{Can grounded and verified claims serve as a reliable context for question answering?}
The retrieval experiment in Section~\ref{sec:retrieval} evaluates GRACE as a static knowledge base. However, the core premise of our framework is that the knowledge base should \textit{grow}: expert-verified boundary claims become new priors that expand coverage and improve downstream performance over successive iterations. 
This experiment tests whether the co-evolutionary loop produces a knowledge base that serves as increasingly reliable context for question answering. 

\paragraph{Setup.} We evaluate the knowledge expansion loop on 100 QASPER papers (334 questions total), using pre-built bipartite graphs initialized with Abstract + Introduction priors (Round 0). 
In each of three expansion rounds (Round 1-3), a responder LLM (GPT-4o-mini) generates answers using a progressively revealed batch of held-out paper sections as out-of-distribution (OOD) context. The responses are decomposed into claims and verified by an LLM verifier (GPT-4.1) that has the full paper as context. 
We use the LLM verifier as a proxy for human experts, allowing fully automated evaluation of the expansion loop. 
Because this LLM verifier setting underpins every result in this section, we also test with human verifiers to compare alignment. A separate study replaces the verifier with 15 graduate students familiar with NLP topics on a 15 paper subset, reported at the end of this section (Table~\ref{tab:human_verification}).
We set $Cost(c)$ as the minimum embedding distance between claim $c$ and the nearest prior, reflecting the assumption that claims most distant from established knowledge are hardest to validate. For human validation, we also add time for validation per claim to the cost function.
Verified claims are promoted to Grounded status with anchor-weight edges ($w = 3.0$), and a cascade re-check propagates grounding to additional Boundary claims that are entailed by the newly verified claim. For downstream QA evaluation, the accumulated Grounded claims at each round serve as the sole context provided to the responder LLM. We compare against an abstract-only baseline and a full-paper upper bound, where the entire paper text is given as context.

\paragraph{Metrics.}
In addition to QA accuracy, we track four metrics across expansion rounds.
\textbf{Prior Coverage Rate (PCR)} is the fraction of all response claims classified as Grounded,
measuring overall knowledge coverage.
\textbf{Knowledge Frontier (KF)} is the average number of Boundary claims remaining per paper,
indicating how much unresolved knowledge persists.
\textbf{Verification Yield} is the fraction of oracle-routed OOD claims confirmed as correct,
measuring the efficiency of the claim selection algorithm.
\textbf{Cascade} is the number of Boundary claims that flipped to Grounded \emph{without}
being directly verified, capturing the indirect grounding propagation triggered by each
verification round.

\paragraph{Results.}
        

\begin{table}[h]
    \centering
    \begin{tabularx}{\linewidth}{l *{2}{>{\centering\arraybackslash}X}}
    \toprule
    \textbf{Context} & \textbf{Accuracy (\%)} & \textbf{$\Delta$} \\
    \midrule
    Abstract only           & 0.1413 & --- \\
    Round 0 (Abstract + Intro) & 0.2225 & $+8.1$ \\
    Round 1 (Batch 1) & 0.2447 & $+10.3$ \\
    Round 2 (Batch 2) & 0.3716 & $+23.0$ \\
    Round 3 (Batch 3) & 0.4913 & $+35.0$ \\
    \midrule
    Full paper & 0.3360 & $+19.5$ \\
    \bottomrule
    \end{tabularx}
    \caption{\textbf{Downstream QA accuracy improves monotonically with each expansion round.} $\Delta$ is relative to the abstract-only baseline.}
    \label{tab:qa_accuracy}
\end{table}
\begin{figure}
    \centering
    \includegraphics[width=\linewidth]{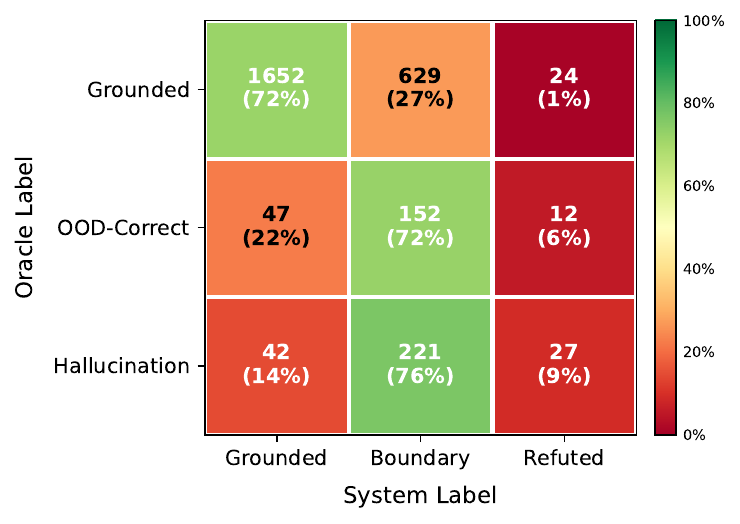}
    \caption{\textbf{GRACE claim classification align with ground truth.} Confusion matrix of system-assigned labels versus oracle labels.}
    \label{fig:confusion}
\end{figure}

\begin{table*}[h]
\centering
\begin{tabularx}{\linewidth}{l l *{4}{>{\centering\arraybackslash}X}}
\toprule
\textbf{Round} & \textbf{Context} & \textbf{PCR$\uparrow$} & \textbf{KF$\downarrow$} & \textbf{Yield} & \textbf{Cascade} \\
\midrule
0 & Abstract + Intro & 63.1\% & 10.02 & --- & --- \\
1 & + Batch 1 & 77.3\% & 6.25 & 56.3\% & 3.77 \\
2 & + Batch 2 & 80.6\% & 5.30 & 58.4\% & 0.95 \\
3 & + Batch 3 & 83.4\% & 4.57 & 69.5\% & 0.73 \\
\bottomrule
\end{tabularx}
\caption{\textbf{Knowledge coverage grows and the boundary shrinks with each expansion round.} PCR rises while verification yield increases across rounds.}
\label{tab:expansion_dynamics}
\end{table*}
Table~\ref{tab:qa_accuracy} reports downstream QA accuracy across expansion rounds.
Starting from 14.1\% with abstract-only context, accuracy rises monotonically to 49.1\% after three rounds of expansion. Notably, Round 3 accuracy surpasses the full-paper oracle (33.6\%) by 15.5 percentage points, demonstrating that a compact pool of verified claims provides more effective QA context than raw full-text, which contains noise from unrelated sections.
Table~\ref{tab:expansion_dynamics} shows the underlying knowledge graph dynamics: PCR grows from 63.1\% to 83.4\%, while KF shrinks from 10.0 to 4.6 Boundary claims per paper.
Verification yield increases across rounds ($56.3\% \rightarrow 69.5\%$), indicating that later batches (covering results and conclusions) produce claims that are more readily verifiable.
Cascade propagation is frontloaded (3.77 claims/paper in Round 1, decaying to 0.73 by Round~3), reflecting the dense mutual entailment structure of methods sections.
Additionally, the confusion matrix in Figure~\ref{fig:confusion} confirms the 
classifier's safety properties: grounding precision reaches 97.6\%, meaning contamination of the Grounded pool by hallucinations is rare (42 out of 1,741 system-Grounded claims, or 2.4\%). The system's primary failure mode is conservative---22.3\% of OOD-correct claims are prematurely promoted to Grounded because they closely paraphrase the abstract---while 85.5\% of hallucinated claims are correctly classified as Boundary or Refuted.
These results demonstrate that GRACE's co-evolutionary loop produces a knowledge base that is both safer and more effective than raw document context, with each verification round compounding coverage gains through cascade propagation.
\paragraph{Human Verification.}
\begin{table*}[h]
\centering
\begin{tabularx}{\linewidth}{l *{5}{>{\centering\arraybackslash}X}}
\toprule
\textbf{Round} & \textbf{Accuracy} & \textbf{PCR$\uparrow$} & \textbf{KF$\downarrow$} & \textbf{Yield} & \textbf{Abstention} \\
\midrule
Round 0 & 16.67\% & 89.55\% & 5.33 & --- & --- \\
Round 1 (LLM validator) & 18.75\% & 93.58\% & 2.93 & 34.48\% & 44.83\% \\
Round 1 (Human) & 18.75\% & 93.63\% & 2.07 & 37.93\% & 3.45\% \\
\bottomrule
\end{tabularx}
\caption{\textbf{Human validators match the LLM judge on knowledge coverage but shrink the knowledge frontier substantially further.} One expansion round on a 15-paper QASPER subset, verified by 15 human experts compared with LLM validators.}
\label{tab:human_verification}
\end{table*}
The verifier in the experiment above is an LLM agent. To our method's validity with human experts, we re-ran one expansion round on a 15-paper subset with the LLM verifier replaced by 15 NLP graduate students, who produced 354 annotations covering 43 Boundary claims at five or more raters overlapped per claim.
Table~\ref{tab:human_verification} shows that the two verifiers are interchangeable on coverage and downstream accuracy but not on how much of the frontier they resolve. 
PCR is almost equivalent (93.63\% vs 93.58\%), and QA accuracy is identical on all 48 questions. For the knowledge frontier (KF), human verification leaves 2.07 Boundary claims per paper against the LLM's 2.93, a gap of 0.87 claims.
This is due to the high rate of abstention from the LLM verifier: given the same 29 claims, the judge returns 10 confirmations, 6 rejections and 13 abstentions, while the human validators return 11, 17 and 1. An abstention leaves a claim on the frontier whereas a rejection removes it, so the judge's 44.83\% abstention rate against the validators' 3.45\% accounts for the KF gap. We interpret this as a difference in decisiveness rather than in accuracy: the study has no gold key against which either verifier can be scored, so Yield is an acceptance rate and not a correctness measure. Overall, we find that human validators bring similar accuracy benefits to the LLM ones, but expand the knowledge base even more. 

\subsection{Scalability}\label{sec:scalability}
\begin{figure}[h]
    \centering
    \includegraphics[width=\linewidth]{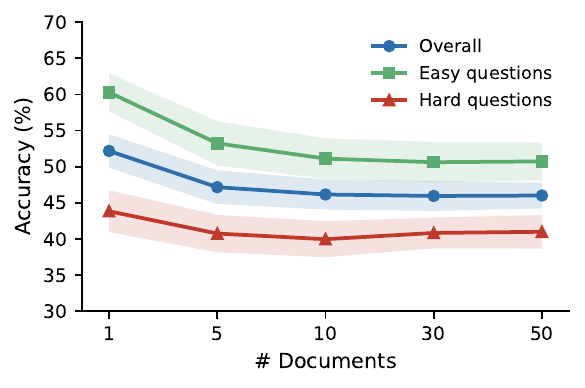}
    \caption{\textbf{GRACE displays reliable retrieval performance on large knowledge bases with multiple long documents.} QA accuracy drops initially at $1\rightarrow 5$ documents, but plateaus when more documents are added.}
    \label{fig:scalability}
\end{figure}

This experiment extends Section~\ref{sec:retrieval} to measure whether GRACE can reliably retrieve relevant context as the knowledge base grows. RAG systems are known to degrade as the number of indexed documents increases, since retrievers must distinguish relevant passages from a larger pool of semantically similar but irrelevant candidates~\cite{shi2023distracted, cuconasu2024power}.
This is a realistic challenge in enterprise settings where knowledge bases span hundreds of documents~\cite{edge2024local}.

We reuse the 50 QuALITY documents and questions from Section~\ref{sec:retrieval}. Instead of retrieving from a single reference document, we add multiple documents to the knowledge graph (5--50), increasing the claim pool from an average of 91 claims per single-article graph to 4{,}554 claims at 50 documents. For all settings, $k=20$ highest-scoring claims are retrieved and passed as numbered context to the responder (GPT-4o-mini), which selects among four multiple-choice options.

\textbf{Results.} As shown in Figure~\ref{fig:scalability}, GRACE maintains largely stable performance as the knowledge base scales from 1 to 50 documents. The largest accuracy drop occurs at the 1$\to$5 document transition ($52.18\%\to47.16\%$, $-5.02$pp), with minimal further degradation beyond that point. Easy questions degrade more than hard ones, likely because precise factual recall is more sensitive to retrieval noise than multi-step reasoning. Overall, despite the claim pool growing $50\times$, accuracy falls by less than 6 percentage points and stabilizes well before the largest pool sizes are reached.

\section{Conclusion}
We introduce GRACE, a graph-grounded framework that addresses LLM hallucination at the systems level by structuring claims and trusted knowledge priors into a weighted bipartite graph. Through weighted closeness centrality analysis, GRACE maps the LLM's knowledge frontier and enables separation of verified facts from hallucinations and novel knowledge. The Return on Attention framework allocates scarce expert verification effort by only routing claims whose utility exceeds the cost of review. 
Our experiments demonstrate that GRACE outperforms RAG-based baselines as a retrieval tool, that the ROA-guided claim selection efficiently identifies high-value boundary knowledge for expansion, and that the framework scales reliably as the underlying corpus grows.
In future work, we plan to integrate real-world data from various use cases in the industry to further validate GRACE's utility in expert-in-the-loop knowledge expansion.

\section*{Limitations}
Three limitations bound this work. First, graph construction is the dominant cost. Claim decomposition and pairwise prior-claim entailment sweep rely on LLM calls, and it is reducible by replacing LLM decomposition with rule-based extraction for structured priors and by pre-filtering the entailment sweep so that only ambiguous pairs reach the LLM. 
Second, our human verification study is small: one expansion round with 15 participants. Time and resources allowing, we plan to recruit more participants to validate claims from more sources, and expand to more rounds to measure GRACE's function on multiple human-in-the-loop expansion iterations.
Lastly, GRACE assumes every prior is equally trustworthy, giving all prior-claim edges the same weight $w_{\text{anchor}}$. However, real sources go out of date, are superseded, or are uncertain in themselves. In future developments, we plan to introduce a per-prior trust term derived from provenance, recency, and expert reliability, propagated through $CC(v)$ so that grounding reflects confidence in the priors and not only proximity to them.

\section*{Acknowledgments}
This project was supported by a grant from MassMutual.


\bibliography{custom}


\end{document}